\documentclass[11pt]{article}

\usepackage[utf8]{inputenc}
\usepackage[T1]{fontenc}
\usepackage{times}
\usepackage{geometry}
\usepackage{amsmath, amssymb}
\usepackage{graphicx}
\usepackage{enumitem}
\usepackage{url}
\usepackage{natbib}
\usepackage{titlesec}
\usepackage{hyperref}
\usepackage{xcolor}
\usepackage{booktabs}
\hypersetup{
    colorlinks=true,
    linkcolor=blue,
    citecolor=blue,
    urlcolor=blue
}

\titleformat{\section}[block]{\normalfont\Large\bfseries}{\thesection.}{1em}{}
\titleformat{\subsection}[block]{\normalfont\large\bfseries}{\thesubsection.}{1em}{}

\title{Order from Chaos: Comparative Study of Ten Leading LLMs on Unstructured Data Categorization}
\author{
  Ariel Kamen \\
  RingCentral, UC Davis \\
  \texttt{ariel.kamen@ringcentral.com}
}
\date{Sep. 15, 2025}

\begin{document}

\maketitle

\begin{abstract}
This study presents a comparative evaluation of ten state-of-the-art large language models (LLMs) applied to unstructured text categorization using the Interactive Advertising Bureau (IAB) 2.2 hierarchical taxonomy. The analysis employed a uniform dataset of 8,660 human-annotated samples and identical zero-shot prompts to ensure methodological consistency across all models. Evaluation metrics included four classic measures—accuracy, precision, recall, and F1-score—and three LLM-specific indicators: hallucination ratio, inflation ratio, and categorization cost.

Results show that, despite their rapid advancement, contemporary LLMs achieve only moderate classic performance, with average scores of 34\% accuracy, 42\% precision, 45\% recall, and 41\% F1-score. Hallucination and inflation ratios reveal that models frequently overproduce categories relative to human annotators. Among the evaluated systems, Gemini 1.5/2.0 Flash and GPT 20B/120B offered the most favorable cost-to-performance balance, while GPT 120B demonstrated the lowest hallucination ratio. The findings suggest that scaling and architectural improvements alone do not ensure better categorization accuracy, as the task requires compressing rich unstructured text into a limited taxonomy—a process that challenges current model architectures.

To address these limitations, a separate ensemble-based approach was developed and tested. The ensemble method, in which multiple LLMs act as independent experts, substantially improved accuracy, reduced inflation, and completely eliminated hallucinations. These results indicate that coordinated orchestration of models—rather than sheer scale—may represent the most effective path toward achieving or surpassing human-expert performance in large-scale text categorization

\end{abstract}

\vspace{1em}

\section{Introduction}

Text categorization is a core task in natural language processing (NLP), supporting applications such as spam filtering, sentiment analysis, document retrieval, and content moderation. Early approaches relied on manual expert annotation, rule-based heuristics, or traditional machine learning models that required domain-specific training and extensive feature engineering. While effective, these methods often demanded significant resources and lacked scalability.

Recent advances in large language models (LLMs) such as OpenAI’s GPT, Google’s Gemini, Anthropic’s Claude, xAI’s Grok, Meta’s LLaMA, Mistral, and DeepSeek have introduced strong zero-shot classification capabilities, lowering the barrier to deploying text categorization systems. Yet their application raises key questions: how do LLMs compare to traditional models, what hidden costs and limitations emerge, and which models provide the best trade-off between quality and efficiency? To address these questions, we benchmark ten major LLMs on an 8,660-document corpus annotated with the Interactive Advertising Bureau (IAB 2.2) taxonomy, evaluating both standard metrics (accuracy, precision, recall, F1-score) and three LLM-specific measures: hallucination ratio, category inflation ratio, and token-processing cost. We further examine the impact of prompt design and API-level hyperparameter variations (temperature, top-k, and maximum tokens), offering a comprehensive assessment of the strengths, limitations, and practical trade-offs of LLMs in large-scale unstructured text categorization.

The next section reviews related work in traditional text classification methods and recent studies of LLMs applied to categorization tasks.

\section{Related Work}

Classification systems and taxonomies are a cornerstone of modern information systems. Human categorization, once an intellectual and interpretive activity, has gradually transitioned into mechanical and computational forms. Traditional approaches relied on matching new content to labeled exemplars or on manually constructed rule-based systems. These methods were widely deployed in industrial contexts such as spam detection, content moderation, and programmatic advertising. While precise in narrow domains, rule-based systems were difficult to scale, struggled to generalize to unstructured documents, and remained heavily language dependent.

The rise of machine learning significantly broadened the scope of text categorization. Models such as logistic regression, support vector machines (SVMs) \citep{joachims1998svmtext}, random forests \citep{breiman2001randomforests}, and shallow neural networks gained popularity due to their ability to generalize from labeled data. Comprehensive surveys such as \citet{Sebastiani2002} document the rapid development of automated text categorization during this period. However, these methods typically required complex feature engineering and large annotated corpora, making them resource intensive. Deep learning approaches \citep{lecun2015deeplearning} reduced the burden of manual feature design but introduced new challenges in training, computation, and deployment.

The introduction of transformer-based architectures and large-scale pretraining, exemplified by BERT \citep{devlin2019bert} and GPT \citep{radford2019lmmultitask}, transformed the field. These models, and more recently large language models (LLMs), demonstrated unprecedented ability to generalize to unseen tasks, including text classification. Nevertheless, concerns remain: \citet{xu-etal-2024-classification} note that LLMs may suffer from benchmark contamination and overfitting, particularly when evaluation datasets overlap with pretraining distributions. Despite these limitations, their strong zero-shot and few-shot performance has spurred widespread interest in applying LLMs to real-world categorization.

To address inherent weaknesses, several frameworks have proposed structured prompting and model decomposition. For example, the CARP framework \citet{sun-etal-2023-carp} decomposes classification into simpler subtasks to improve performance in hierarchical taxonomies. SPIN \cite{jiao-etal-2024-spin} prunes internal neurons to emphasize task-relevant features, while \citet{edwards-etal-2024-incontext} investigate in-context learning for text classification, finding that results vary substantially with taxonomy depth and complexity.

Although most prior work has focused on accuracy and overall classification performance, relatively few studies have examined zero-shot LLMs as a dedicated solution for categorization and thus have not addressed LLM-specific factors such as computational cost, hallucination, or category inflation. To our knowledge, no prior study has applied LLMs to the Interactive Advertising Bureau (IAB) taxonomy, despite its status as one of the most widely used categorization frameworks in internet marketing and advertising. This study addresses that gap by systematically evaluating ten major LLMs on an IAB-based categorization task, incorporating both traditional metrics and novel measures of real-world relevance.

\section{Methodology}

\subsection{Dataset Compilation}

The dataset consists of 8,660 unstructured textual samples drawn from diverse topical domains. The texts were sourced from open news corpora and manually categorized by expert annotators using the 690 general-purpose categories of the IAB 2.2 taxonomy \citep{kamen2025unstructureddataset}. Each sample was assigned one or more categories judged by human experts to best represent its content. Owing to the hierarchical structure of the IAB taxonomy, these categories may be drawn from different tiers of the taxonomy.

\subsection{LLM Selection and Configuration}

We evaluated ten publicly available and widely used LLMs: Anthropic’s Claude 3.5; Google’s Gemini 1.5 and Gemini 2.0 Flash; Meta’s LLaMA 3.3 70B and LLaMA 3 8B; Mistral’s Mistral-Large-Latest Nano; xAI’s Grok; Groq-hosted DeepSeek’s DeepSeek; and Groq-hosted GPT OSS-20B and GPT OSS-120B. All models were accessed through their official APIs directly or using  Groq hosting services. To ensure fair comparison, evaluations were conducted on the same dataset using a same prompting strategy.

\subsection{Categorization Schema}

Categorization follows the Interactive Advertising Bureau (IAB) taxonomy, version 2.2 \citep{IAB2022}. The IAB framework functions as a de-facto industry standard in online advertising and internet marketing and is used by hundreds of internet publishers, advertisers, and technology providers. The taxonomy contains 690 general-purpose categories and subcategories organized in a four-tier hierarchy. We restrict our experiments to the general-purpose portion of taxonomy. Figure 1 illustrates the IAB hierarchical structure.

\subsection{Prompting Procedure}

 We employ an iterative, hierarchy-aware prompting strategy that mirrors human taxonomy navigation. The model first selects a Tier-1 category, then progressively refines its choice through Tier-2, Tier-3, and Tier-4 prompts. To minimize hallucinations and ensure validity, taxonomy constraints such as allowable children, canonical identifiers, and formatting requirements are embedded directly into the prompts. All outputs are normalized to canonical IAB labels before scoring. The exact prompt template used in the experiments is provided below.

\begin{quote}
    Your job is to categorize unstructured text according to the following list of categories. You will be given a certain amount of text from the text. Your response should contain only the categories, with no other text. If the text fits multiple categories, output them separated by a comma and a space. Categories are separated via comma. You may not output categories not in the list. If no categories fit the text, output 'None'. Categories: {categories}"
\end{quote}

\subsection{Hyperparameter Sweeps}

To test sensitivity to decoding settings, we implemented a parameter-optimization harness over temperature, top-k, and maximum tokens. We conducted thousands of runs across models and tiers. Consistent with the short, schema-constrained nature of the outputs, hyperparameter variation produced negligible changes in categorization quality, while occasionally affecting verbosity or formatting. We therefore report main results under standardized defaults.

\subsection{Evaluation Protocol (Overview)}

Performance is measured on the 8,660-document benchmark using accuracy, precision, recall, and F1-score. We further compute three LLM-specific measures introduced in this work: hallucination ratio, category inflation ratio, and token-processing cost. Formal definitions and matching rules  are detailed in Section 4 (Categorization Models and Evaluation Criteria). All invalid or out-of-taxonomy predictions are counted toward hallucination; multi-label emissions where a single label is required contribute to inflation. 

\begin{figure}[h]
    \centering
    \includegraphics[width=0.9\linewidth]{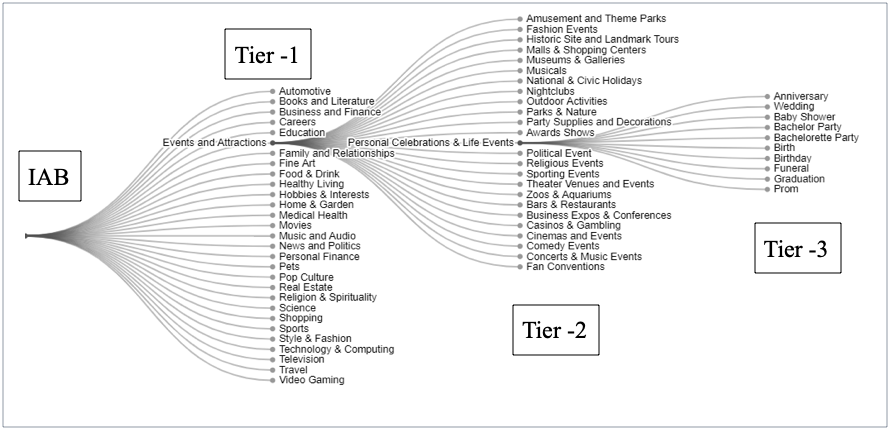}
    \caption{Overview of the IAB Taxonomy}
    \label{fig:iab-taxonomy}
\end{figure}

\section{Categorization Models and Evaluation Criteria}

Let T denote a hierarchical taxonomy of categories.

The categorization of a text x is defined as generating a set of categories that describe its content:

\begin{equation}
\mathrm{LLM}_{T,x}=\{t_1,t_2,\ldots,t_n\}
\tag{4.1}
\end{equation}

Each predicted category may either belong to the taxonomy T or represent a hallucination (i.e., a category outside T).

For evaluation, we assume access to an expert (human or other reliable ground truth source) that categorizes the same text:

\begin{equation}
E(T,x)=\{e_1,e_2,\ldots,e_m\},\quad e_j\in T
\tag{4.2}
\end{equation}

We define:
\begin{itemize}
    \item True Positives (TP): categories assigned by both the LLM and the expert.
    \item False Positives (FP): categories predicted by the LLM but absent in the expert set.
    \item False Negatives (FN): categories assigned by the expert but missed by the LLM.
\end{itemize}

Using these, the classic evaluation criteria are:

\subsection{Classic Criteria}

\begin{itemize}
  \item Accuracy: overall proportion of correct predictions.
  \begin{equation}
  \mathrm{Accuracy}_x=\frac{TP_x}{TP_x+FP_x+FN_x}
  \tag{4.3}
  \end{equation}

  \item Precision: fraction of predicted categories that are correct.
  \begin{equation}
  \mathrm{Precision}_x=\frac{TP_x}{TP_x+FP_x}
  \tag{4.4}
  \end{equation}

  \item Recall: fraction of expert categories recovered by the model.
  \begin{equation}
  \mathrm{Recall}_x=\frac{TP_x}{TP_x+FN_x}
  \tag{4.5}
  \end{equation}

  \item F1-score: harmonic mean of Precision and Recall, balancing both metrics into a single measure.
  \begin{equation}
  \mathrm{F1}_x=\frac{2\,\mathrm{Precision}_x\,\mathrm{Recall}_x}{\mathrm{Precision}_x+\mathrm{Recall}_x}
  \tag{4.6}
  \end{equation}
\end{itemize}

\subsection{Additional Criteria for LLM-Based Categorization}

To capture behaviors specific to LLMs, we introduce three additional metrics:
\begin{itemize}
  \item Hallucination Ratio (HR): measures how often the LLM produces categories outside the taxonomy.
  \begin{equation}
  h_x=\{\,t_i\in \mathrm{LLM}_{T,x}\;:\; t_i\notin T\,\}
  \tag{4.7}
  \end{equation}
  \begin{equation}
  \mathrm{HR}_x=\frac{|h_x|}{|\mathrm{LLM}_{T,x}|}
  \tag{4.8}
  \end{equation}

  \item Category Inflation Ratio (IR): compares the number of predicted categories with the number assigned by the expert.
  \begin{equation}
  \mathrm{IR}_x=\frac{|\mathrm{LLM}_{T,x}|}{|E(T,x)|}
  \tag{4.9}
  \end{equation}
  \end{itemize}
  To avoid redundancy in hierarchical taxonomies, we apply the Parent Exclusion Rule (PER), which removes a parent if its child is present. The reduced inflation ratio is:
  \begin{equation}
  \mathrm{IR}^{*}_x=\frac{|\mathrm{PER}(\mathrm{LLM}_{T,x})|}{|E(T,x)|}
  \tag{4.10}
  \end{equation}
\begin{itemize}
  \item Price of Computation (Cost): quantifies monetary expense based on token usage.
  \begin{equation}
  \mathrm{Cost}_x=c_{\mathrm{in}}\,N^{(x)}_{\mathrm{in}}+c_{\mathrm{out}}\,N^{(x)}_{\mathrm{out}}
  \tag{4.11}
  \end{equation}
  \end {itemize}
  where $c_{\mathrm{in}}$ and $c_{\mathrm{out}}$ are the prices per input and output token, and $N^{(x)}_{\mathrm{in}}$, $N^{(x)}_{\mathrm{out}}$ are the respective input and output token counts.

\subsection{Corpus-level Aggregation}

For the corpus of texts $D=\{x_1,\ldots,x_k\}$, metrics are aggregated as:
\begin{equation}
\mathrm{Metric}^{\mathrm{macro}}(D)=\frac{1}{k}\sum_{i=1}^{k}\mathrm{Metric}(x_i)
\tag{4.12}
\end{equation}

The total computation cost is:
\begin{equation}
\mathrm{Cost}_{D}=\sum_{i=1}^{k}\mathrm{Cost}(x_i)
\tag{4.13}
\end{equation}

Traditional metrics such as accuracy, precision, recall, and F1-score provide a well-established way to measure correctness. However, LLMs introduce unique challenges: they may generate categories not present in the taxonomy (hallucinations), assign too many or too few labels (inflation or under-assignment), and incur significant computational cost. By incorporating hallucination ratio, inflation ratio, and cost, we capture aspects of performance that are especially relevant to large-scale, real-world deployments of LLM-based categorization. This combination of classical and novel measures allows us to evaluate not only how correct the predictions are, but also how reliable, efficient, and usable the categorization is in practice.

Having established the evaluation framework and introduced both classical and LLM-specific performance criteria, we now turn to a comparative analysis of ten leading models.

\section{Performance Evaluation}

We analyzed ten popular zero-shot LLMs using the same 8,660-sample dataset and a unified prompt against a human-annotated benchmark. Comparison metrics include four classic metrics (accuracy, precision, recall, and F1-score) and three LLM-specific metrics (hallucination ratio, inflation ratio, and categorization cost). The Groq API was used to evaluate open-source models LLaMA 3 8B, LLaMA 3 70B, GPT 20B, GPT 120B, and DeepSeek R1. Other models were accessed through their providers’ APIs.
Table \ref{tab:classification} reports corpus-level aggregated accuracy, precision, recall, and F1-score. Classic metric performance varied as follows: 34\% for accuracy; 42\% for precision; 45\% for recall; and 41\% for F1-score. Claude 3.5 and the GPT models delivered the strongest classic-metric performance, whereas LLaMA 3 8B and Mistral lagged behind.

\begin{table}[h]
\centering
\caption{LLMs Mean Performance Scores (Sample Size: 8,660)}
\label{tab:classification}
\begin{tabular}{lcccc}
\toprule
\textbf{Model} & \textbf{F1} & \textbf{Accuracy} & \textbf{Precision} & \textbf{Recall} \\
\midrule
Claude 3.5      & 0.55 & 0.52 & 0.46 & 0.79 \\
Gemini 1.5 Flash     & 0.49 & 0.54 & 0.45 & 0.64 \\
Gemini 2.0 Flash& 0.52 & 0.54 & 0.46 & 0.72 \\
LLaMA 3 8B      & 0.39 & 0.41 & 0.33 & 0.60 \\
LLaMA 3.3 70B   & 0.51 & 0.43 & 0.40 & 0.87 \\
DeepSeek        & 0.52 & 0.51 & 0.45 & 0.75 \\
Grok            & 0.50 & 0.55 & 0.46 & 0.66 \\
Mistral         & 0.47 & 0.41 & 0.36 & 0.83 \\
GPT-20B         & 0.52 & 0.55 & 0.47 & 0.71 \\
GPT-120B        & 0.53 & 0.55 & 0.47 & 0.72 \\
\bottomrule
\end{tabular}
\end{table}

Table \ref{tab:pricing} lists each model’s input and output token costs per 1 million tokens (public rates as of September 2025). As expected, open-source token pricing was significantly lower than private models. Input costs ranged from \$0.05 for LLaMA 3 8B to \$0.59 for LLaMA 3 70B per 1 million input tokens. Private LLM input pricing ranged from \$0.80 for Claude 3.5 up to \$8 for Mistral. Gemini 1.5/2.0 Flash, LLaMA 3 8B, and GPT 20B/120B were the most cost-efficient models in this experiment. 

\begin{table}[h]
\centering
\caption{LLM Pricing Models as of September 2025 (per 1M tokens)}
\label{tab:pricing}
\begin{tabular}{lcc}
\toprule
Model & Input Cost & Output Cost \\
\midrule
Claude 3.5            & \$0.80  & \$4.00 \\
Gemini 1.5 Flash      & \$0.075 & \$0.30 \\
Gemini 2.0 Flash      & \$0.10  & \$0.40 \\
Mistral               & \$8.00  & \$8.00 \\
LLaMA 3 8B            & \$0.05  & \$0.08 \\
LLaMA 3 70B           & \$0.59  & \$0.79 \\
Grok                  & \$2.00  & \$10.00 \\
DeepSeek              & \$0.27  & \$1.10 \\
GPT 20B               & \$0.10  & \$0.50 \\
GPT 120B              & \$0.15  & \$0.75 \\
\bottomrule
\end{tabular}
\end{table}

Table \ref{tab:hallucination} summarizes the average categorization cluster size before and after hallucination filtering, the hallucination ratio (HR), and the inflation ratio (IR). Hallucination ratio varies at 843\%, and inflation ratio at 209\%. GPT 120B demonstrated the lowest hallucination, approximately 40\% lower than the next-best Grok. The benchmark dataset averaged 4.01 categories per article, indicating that all LLMs tended to overproduce labels relative to human annotators. Gemini 1.5/2.0, Grok, and GPT 20B/120B showed the lowest inflation. Both LLaMA 3 8B and 70B exhibited high hallucination and inflation, tending to over-generate labels and reduce focus.

\begin{table}[h]
\centering
\caption{Average Categorization Cluster Size and Hallucination Rate}
\label{tab:hallucination}
\begin{tabular}{lccc}
\toprule
\textbf{Model} & \textbf{Avg Cluster Size} & \textbf{Filtered Cluster Size} & \textbf{Hallucination Rate (\%)} \\
\midrule
Claude 3.5           & 6.32 & 6.25 & 1.1\% \\
Gemini 1.5 Flash     & 5.02 & 4.91 & 2.2\% \\
Gemini 2.0 Flash     & 5.73 & 5.61 & 2.1\% \\
LLaMA 3 8B           & 7.08 & 6.71 & 5.4\% \\
LLaMA 3.3 70B        & 10.51 & 9.91 & 5.9\% \\
DeepSeek             & 6.21 & 6.14 & 1.1\% \\
Grok                 & 5.23 & 5.18 & 1.0\% \\
Mistral              & 8.81 & 8.67 & 1.6\% \\
GPT 20B              & 5.41 & 5.34 & 1.3\% \\
GPT 120B             & 5.36 & 5.32 & 0.7\% \\
\bottomrule
\end{tabular}
\end{table}

It is worth noting that hallucination filtering led to only marginal reductions in cluster size (typically less than 0.5 categories per sample) and slight improvements in model performance.

What is the best LLM overall? In price/performance terms, the Gemini and GPT families are the clear winners, and GPT 120B is preferred due to its very low hallucination ratio. 

These observations motivate the following discussion of why classic metrics remain modest for zero-shot categorization, how task structure drives these outcomes, and where model scaling and orchestration strategies might change the picture.

\section{Discussion}

In our comparison, classic performance metrics remain modest. Given the rapid advances and increasing power of contemporary LLMs, one might expect substantially higher accuracy, precision, recall, and F1-scores. Instead, the zero-shot setup produced only moderate results, revealing a clear gap between general generative competence and targeted classification of arbitrary unstructured text within a complex hierarchical taxonomy.

Task specificity represents the primary bottleneck. Categorization requires compressing arbitrary, unstructured texts into a sparse, predefined label space. Although the IAB taxonomy is a well-designed and widely adopted framework, it includes only 690 generic categories—insufficient to reflect the full complexity and diversity of real-world content. This limitation becomes more apparent when compared to the DMOZ taxonomy (the Open Directory Project), which contains over 750,000 nodes spanning numerous hierarchical levels \citep{kosmopoulos2015lshc, hoek2021dmoz, researchgate2010dmozviz}. Achieving human-expert-level IAB categorization therefore demands not only linguistic fluency but also a stable, taxonomy-aware world model capable of mapping dispersed semantic evidence into concise, non-overlapping labels. Standard next-token prediction pretraining does not guarantee this structured compression ability, particularly in zero-shot conditions.

Scaling and architectural improvements alone do not ensure performance gains. Newer or larger models—such as Gemini 1.5/2.0 and GPT 20B/120B—did not consistently outperform their predecessors across all classic metrics. We hypothesize that for categorization tasks, performance improvements diminish beyond a certain scale threshold; architectural refinement, instruction tuning, and prompt design likely play a greater role than raw model size.

Cost also significantly affects the practical usability of LLMs for categorization. Models such as Gemini 1.5/2.0, GPT 20B/120B, and DeepSeek combine low token pricing with competitive accuracy, enabling broader experimentation and deployment. When coupled with a low hallucination ratio, GPT 120B emerges as the strongest overall choice for balanced cost and performance. Hallucination filtering, however, provides only marginal benefits, typically reducing cluster size by less than half a category per sample, indicating that post-processing alone cannot substantially improve precision.

To address these limitations, we explored a different paradigm: LLM ensembles. In this approach, multiple models act as independent experts and make categorization decisions collaboratively. The ensemble framework, tested and optimized in a separate study, substantially improved overall categorization performance, completely eliminated hallucinations, and reduced category inflation. While a detailed analysis of ensemble design lies beyond the scope of this paper, these findings suggest that orchestration—rather than scale alone—offers a promising path toward achieving or surpassing human-expert consistency in large-scale text categorization.

\section{Conclusion}

Zero-shot LLMs can perform large-scale text categorization with consistent and reproducible behavior; however, their classic performance metrics remain modest, and their outputs tend to inflate category sets relative to human annotations. Cost-efficient models such as Gemini 1.5/2.0 and GPT 20B/120B offer strong price-to-performance profiles, with GPT 120B standing out for its notably low hallucination ratio. The core challenge remains structural: compressing semantically rich text into a sparse taxonomy demands disciplined, taxonomy-aware reasoning that current general-purpose pretraining does not yet guarantee.

Near-term progress is most likely to come from orchestration rather than scale. The collaborative use of multiple LLMs organized in an ensemble has proven to significantly improve categorization quality while maintaining efficiency. The additional computational cost of ensemble-based categorization appears to be an attractive trade-off for its multiple benefits, complete elimination of hallucinations, reduced category inflation, and a consistent level of performance that may match or even surpass human experts.

\nocite{GoogleGemini2025}
\nocite{AnthropicClaude2025}
\nocite{MetaLlama2025}
\nocite{XAiGrok2025}
\nocite{DeepSeekChat2025}
\nocite{MistralAI2025}

\bibliographystyle{plainnat}
\bibliography{references}

\end{document}